\documentclass[11pt,a4paper]{article}
\usepackage[hyperref]{acl2020}
\usepackage{times}
\usepackage{latexsym}

\usepackage{graphicx}

\usepackage{microtype}

\aclfinalcopy 

\title{Improving Truthfulness of Headline Generation}

\author{Kazuki Matsumaru \\
 \\\And
  Sho Takase \\
  \{kazuki.matsumaru, sho.takase, naoaki.okazaki\}@nlp.c.titech.ac.jp \\
  Tokyo Institute of Technology \\ \And
  Naoaki Okazaki \\
   \\}

\date{}

\begin{document}
\maketitle
\begin{abstract}
Most studies on abstractive summarization report ROUGE scores between system and reference summaries.
However, we have a concern about the {\it truthfulness} of generated summaries: whether all facts of a generated summary are mentioned in the source text.
This paper explores improving the truthfulness in headline generation on two popular datasets.
Analyzing headlines generated by the state-of-the-art encoder-decoder model, we show that the model sometimes generates untruthful headlines.
We conjecture that one of the reasons lies in untruthful supervision data used for training the model.
In order to quantify the truthfulness of article-headline pairs, we consider the textual entailment of whether an article entails its headline.
After confirming quite a few untruthful instances in the datasets, this study hypothesizes that removing untruthful instances from the supervision data may remedy the problem of the untruthful behaviors of the model.
Building a binary classifier that predicts an entailment relation between an article and its headline, we filter out untruthful instances from the supervision data.
Experimental results demonstrate that the headline generation model trained on filtered supervision data shows no clear difference in ROUGE scores but remarkable improvements in automatic and manual evaluations of the generated headlines.
\end{abstract}

\section{Introduction}
Automatic text summarization aims at condensing a text into a shorter version while maintaining the essential information~\cite{mani:01}.
Methods on summarization are broadly categorized into two approaches: {\it extractive} and {\it abstractive}.
The former extracts important words, phrases, or sentences from a source text to compile a summary~\cite{Goldstein:2000,Erkan:2004,Mihalcea:04,Lin:2011}.
In contrast, the latter involves more complex linguistic operations (e.g., abstraction, paraphrasing, and compression) to generate a new text~\cite{Knight:2000,Clarke:2008}.
Until 2014, abstractive summarization had been less popular than extractive one because of the difficulty of generating a natural text.
However, research on abstractive summarization has attracted a lot of attentions recently with the advances on encoder-decoder models~\citep{rush-etal-2015-neural,takase-etal-2016-neural,zhou-etal-2017-selective,cao-etal-2018-retrieve,song2019mass,wang-etal-2019-biset}. 

English Gigaword \citep{graff2003english,napoles-etal-2012-annotated} is a representative dataset for abstractive summarization.
\citet{rush-etal-2015-neural} regarded Gigaword as a corpus containing a large number of article-headline pairs for training an encoder-decoder model.
Their work assumed a task setting where the first sentence of an article is a source text and its corresponding headline is a target text (summary).
Since then, it has been a common practice to use the Gigaword dataset with this task setting and to measure the quality of generated headlines with ROUGE scores~\cite{Lin:2003} between system-generated and reference headlines.

Apparently, a summarization method is desirable to achieve a ROUGE score of $100$, i.e., a system output is identical to the reference.
However, this is an unrealistic goal for the task setting on the Gigaword dataset. 
The summarization task is underconstrained in that the importance of a piece of information highly depends on the expectations and prior knowledge of a reader~\cite{kryscinski-etal-2019-neural}.
In addition, the Gigaword dataset (as well as other widely-used datasets) was noisy for summarization research because it was not created for the research objective but other professional activities (e.g., news production and distribution).
Thus, the state-of-the-art method could only reach ROUGE-1 scores less than $40$ on the dataset.

While a number of methods compete with each other for the underconstrained task on the noisy data, we have another concern about the {\it truthfulness} of generated summaries: whether all facts of a generated summary are mentioned in the source text.
Unlike extractive summarization, abstractive summarization has no guarantee of truthfulness.
This may result in a serious concern of practical applications of abstractive summarization when a generated summary includes fake facts that are not mentioned in the source document.

In this paper, we explore improving the truthfulness in abstractive summarization on two datasets, English Gigaword and JApanese MUlti-Length Headline Corpus (JAMUL)~\cite{Hitomi2019ALM}.
In Section \ref{sec:unexpected-outputs}, we analyze headlines generated by the state-of-the-art encoder-decoder model and show that the model sometimes generates unexpected words.
In order to estimate the truthfulness to the original text, we measure the recall-oriented ROUGE-1 scores between the source text and the generated headlines.
This analysis reveals that a high ROUGE score between a reference and headline does not necessarily mean a high truthfulness to the source and that there is only a weak correlation between the two.

In Section \ref{sec:analysis-dataset}, we conjecture that one of the reasons why the model sometimes exhibits such an untruthful behavior lies in untruthful article-headline pairs, which are used for training the model.
In order to quantify the truthfulness of article-headline pairs, we consider the textual entailment of whether an article (source document) entails its headline.
We will show that about 30--40\% of source documents do not entail their headlines under the widely-used experimental settings.
In other words, the current task setting is inappropriate for abstractive summarization.
We release the annotations of textual entailment for both English Gigaword and JAMUL\footnote{\url{https://github.com/nlp-titech/headline-entailment}}.

After confirming the untruthfulness of article-headline pairs in the datasets, we hypothesize that removing untruthful instances from the training data may remedy the problem of the untruthful behavior of the model.
In Section \ref{sec:improve-data}, we build a binary classifier that predicts an entailment relation between an article and its headline and use the classifier to filter out untruthful instances in the training data.
We train a model on the filtered supervision data in Section \ref{sec:train}.
Experimental results demonstrate that the filtering procedure shows no clear difference in ROUGE scores but remarkable improvements when we manually and automatically evaluate the truthfulness of the generated headlines.
These results suggest the importance of evaluating truthfulness in addition to relevance.

\section{Unexpected outputs}\label{sec:unexpected-outputs}

\begin{table*}
\hbox to\hsize{\hfil
\begin{tabular}{l|p{9cm}|p{5cm}}\hline
    \multicolumn{1}{c|}{\textbf{\#}} &
    \multicolumn{1}{c|}{\textbf{Input (lead sentence)}} & \multicolumn{1}{c}{\textbf{Output (generated headline)}}  \\ \hline
    1 &
u.s. home resales posted the largest monthly increase in at least \#\# years last month as first-time buyers rushed to take advantage of a tax credit that expires this fall . & home sales rise \#.\# percent \underline{in} \underline{november} \\ \hline
    2 &
seattle -- for years , the standard treatment for patients with blood clots in veins deep in a limb has been blood thinners that stop the clots from getting bigger . & UNK drug may help treat UNK clots \underline{the detroit news} \\ \hline
    3 &
wigan moved to consolidate their premiership status tuesday by tying down one of the brightest stars of last season 's maiden top flight campaign . & \underline{english football league tables} \\ \hline
    4 &
never mind that she has dark blond hair and light blue eyes and the fairest of skin . & \underline{african-american girl is a UNK} \\ \hline
\end{tabular}\hfil}
\caption{\label{tab:gushed-example}
Examples of unexpected outputs generated by the state-of-the-art model. `\#' stands for a digit mask. `UNK' denotes an out-of-vocabulary word. The underlined parts indicate unexpected words.
} 
\end{table*}

\subsection{Examples of unexpected outputs}

Although the current state-of-the-art method for abstractive summarization could only achieve a ROUGE-1 score of less than $40$ on the Gigaword dataset, generated headlines actually look very fluent. This is probably because the encoder-decoder model acquired a strong language model from the vast amount of supervision data.
However, some studies reported that the generated headlines often deviate from the content of the original document~\citep{Cao2018FaithfulTT,kryscinski-etal-2019-neural}. They addressed the problem where an abstractive model made mistakes in facts (e.g., tuples of subjects, predicates, and objects).

However, we also regularly see examples where the abstractive model generates unexpected words.
This is true even for the state-of-the-art model.

Table~\ref{tab:gushed-example} shows examples of unexpected outputs from UniLM~\citep{unilm}, which shows the highest ROUGE scores\footnote{UniLM model fine-tuned on Gigaword dataset achieved $38.90$ ROUGE-1, $20.05$ ROUGE-2, and $36.00$ ROUGE-L scores as of November 22, 2019.} on English Gigaword.
In the first example, the output includes ``in November'' whereas the input did not mention the exact month.
In fact, this article was published in August 2009; however, the model probably guessed the month from the expression ``this fall''.
The second example also exhibits a similar problem where the model incorrectly supplemented the news source ``the Detroit News''.
The third and fourth examples are more problematic in that the generated headlines do not summarize the input sentences at all.

\subsection{Estimating truthfulness}\label{sec:estimating-truthfulness}

\begin{figure}
    \center
    \includegraphics[width=\hsize]{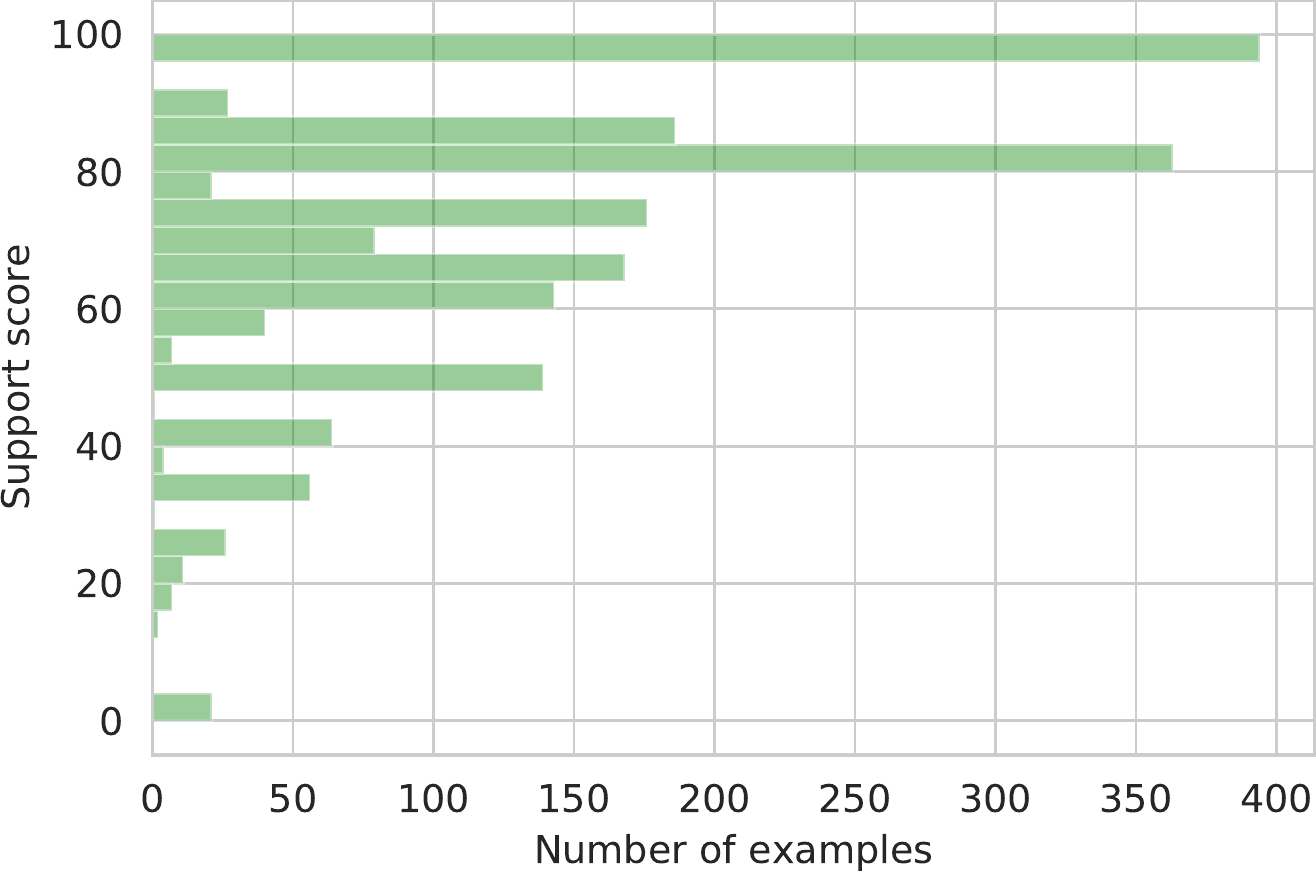}
    \caption{Histogram of support scores (recall-oriented ROUGE-1 scores between generated headlines and their source documents).}
    \label{fig:unilm-rouge-recall}
\end{figure}

In order to quantify the problem of outputs that are untruthful to source documents, we measure the word overlap between the input and output of the UniLM model on the test set of English Gigaword~\cite{rush-etal-2015-neural}.
Here, we calculate the recall-oriented ROUGE-1 score\footnote{We used SumEval:\\ \url{https://github.com/chakki-works/sumeval}}, regarding an output (generated headline) as a {\it gold standard} and an input (source document) as a {\it target to be evaluated}\footnote{We ignore instances whose source documents are less than ten characters long. The total number of instances after this treatment is 1,936.}.
Although this use of the ROUGE metric is unconventional, the intention here is to measure how many words in a generated headline originate from the input document.
In other words, if all words in a generated headline are covered by its source document (truthful), the score is $100$; if none of the words in a generated headline originate from its source document (untruthful), the score is $0$.
We call this ROUGE score {\it support score} hereafter to avoid naming conflicts with conventional ROUGE scores between system and reference summaries.
We mention that we can find a similar method to the support score in several studies; for example, \citet{zhang-etal-2018-abstractiveness} measured the abstractiveness of an output.
Our support score is roughly a reverse version of abstractiveness because the abstractiveness measures the number of words in an output that do not appear in the input.

Figure~\ref{fig:unilm-rouge-recall} reports the histogram of the support scores.
A certain amount of instances receive relatively high support scores: 50.10\% of the instances obtain scores larger than $80$.
At the same time, a non-negligible amount (9.14\%) of instances have support scores less than $40$.
Note that the support scores present rough estimations of the truthfulness of the model; a lower score may imply that a headline includes paraphrased or shortened words from its source document.
Having said that, Figure \ref{fig:unilm-rouge-recall} indicates that the state-of-the-art model sometimes generates untruthful headlines.

\begin{figure}
    \center
    \includegraphics[width=\hsize]{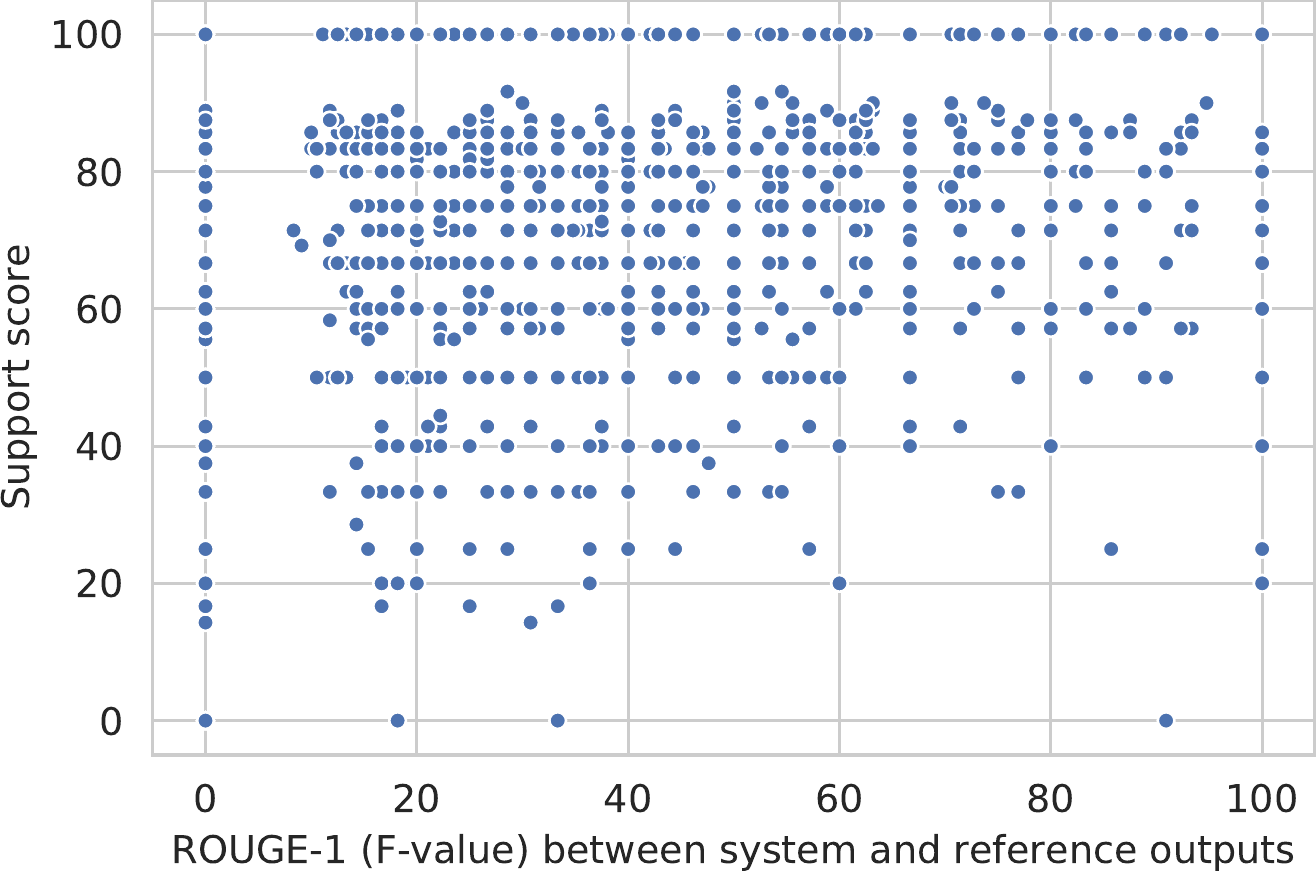}
    \caption{Scatter plots of ROUGE scores and support scores: X-axis presents ROUGE-1 score between system and reference headlines; and Y-axis presents support score (the same to Figure \ref{fig:unilm-rouge-recall}).}
    \label{fig:unilm-rouge-scatter}
\end{figure}

Here, another interesting question comes into our mind: how do the widely-used benchmarking performance values (measured by ROUGE scores between system and reference headlines) reflect the truthfulness (measured by the support scores)?
Figure \ref{fig:unilm-rouge-scatter} depicts the correlation between the two: the X-axis presents the ROUGE-1 score between system and reference headlines, and Y-axis presents support score.
Unfortunately, we cannot observe a strong correlation between the two scores: Pearson's correlation coefficient between the two scores is 0.189, which suggests no correlation.
This result supports that the conventional ROUGE scores tell us little about the truthfulness of generated summaries.

\section{Are the task settings truthful?}\label{sec:analysis-dataset}

\subsection{Background of the datasets and settings}

\begin{table*}
\center
\begin{tabular}{lrrrrr}
\hline
\textbf{data} & \textbf{\# docs} & \textbf{\# words} & \textbf{\# sent / doc} & \textbf{\# words / doc} & \textbf{\# words / headline} \\
\hline
English Gigaword
& 8.6 M &
\begin{tabular}{r}
77 M \\ 4 B
\end{tabular}
& 20.3 & 477.6 & 8.9 \\
\citet{rush-etal-2015-neural} & 3.8 M & 
\begin{tabular}{r}
31 M \\ 119 M
\end{tabular}
& 1 & 31.3 & 8.3 \\
\hline
JAMUL & 1.5 k & 
\begin{tabular}{r}
23 k \\ 547 k
\end{tabular}
& 11.7 & 359.2 & 15.3 \\
JNC & 1.8 M &
\begin{tabular}{r}
26 M \\ 171 M
\end{tabular}
& 3 & 93.2 & 14.2 \\
 \hline
\end{tabular}
\caption{\label{tbl:datasets}
The statistics of datasets and task settings.
The column ``\# words'' presents two values for each row: a top value is the total number of words in the headline; and the bottom value is the total number of words in the article.
The second row of each group (\citet{rush-etal-2015-neural} and JNC) corresponds to the setting of training data.
The columns ``\# sent / doc'', ``\# words / doc'', and ``\# words / headline'' denote the average number of sentences per source document, words per source document, and words per headline, respectively.
}
\end{table*}

Why does a headline generation model exhibit untruthful behavior as we saw in the previous section? Before discussing the reason behind this, we need to understand how the datasets and task settings were established.

The Annotated English Gigaword corpus\footnote{\url{https://catalog.ldc.upenn.edu/LDC2012T21}} is one of the most popular corpora in abstractive summarization research.
\citet{rush-etal-2015-neural} converted this corpus into a dataset for abstractive summarization.
They assumed the lead (first) sentence of an article as a source document and its corresponding headline as a target output.
They did not explain the reason why they did not use a full-length article but only a lead sentence as a source document for headline generation.
We infer that the reason for this treatment is that: a lead sentence provides a strong baseline for extractive summarization; their intention was to explore the capability of abstractive summarization from a lead sentence to a headline; using full text was time-consuming for encoder-decoder models.

Moreover, \citet{rush-etal-2015-neural} introduced some heuristics to remove some noisy instances. They discarded an instance if: (1) the source and target documents have no non-stop word in common; (2) the headline contains a byline or other extraneous editing marks; and (3) a headline includes a question mark or colon.

JApanese MUlti-Length Headline Corpus (JAMUL)\footnote{\url{https://cl.asahi.com/api_data/jnc-jamul-en.html}} is a dataset specially designed for evaluating summarization methods.
JAMUL consists of 1,524 Japanese full-text articles and their print headlines (used for newspapers).
Although JAMUL is distributed for free of charge, JAMUL alone is insufficient for training an encoder-decoder model.
\newcite{Hitomi2019ALM} also released Japanese News Corpus (JNC), which is a large-scale dataset consisting of 1,831,812 pairs of newspaper articles and their print headlines.
JNC includes only the first three sentences of each article\footnote{This is because the price of the dataset would be much higher if it included full-text articles.}.

Table \ref{tbl:datasets} summarizes the datasets and task settings.
As we can see from the rows of \citet{rush-etal-2015-neural} and JNC, these task settings do not use full-text articles but only lead (6.6\% of words in full articles, Gigaword) and lead three sentences (25.9\% of words in full articles, JNC) as source documents for abstractive summarization.
Hence, we hypothesize that the source documents under these task settings contain insufficient information for generating headlines.
In other words, headline generation models might be faced with supervision data where headlines cannot be generated from source documents and learned to be untruthful, i.e., producing pieces of information that are not mentioned in source documents.

\subsection{Truthfulness of the datasets and settings measured by textual entailment}

\begin{table*}
\hbox to\hsize{\hfil
\begin{tabular}{l|p{8cm}|p{4.7cm}|c}\hline
    \textbf{\#} &
    \multicolumn{1}{c|}{\textbf{Source document (text)}} & \multicolumn{1}{c|}{\textbf{Headline (hypothesis)}} & \textbf{Entail} \\ \hline
    1 & 
    France {\it hopes to secure the contract for the supply of Agosta-class submarines} to the Malaysian navy...& France {\it keen to sell submarines} to Malaysian navy & Y \\ \hline
    2 & 
    {\it 69,700 local people to work} & {\it 70,000 employees} & Y \\ \hline
    3 & 
    British boxing promoter Frank Warren on Tuesday announced the signing of three world title contenders. & Three \underline{foreign} boxers join British stable & N \\ \hline
    4 & 
    Lazio and Roma will be playing for more than local bragging rights when they meet...
    & \underline{Football : Italian Serie A table} & N \\ \hline
\end{tabular}\hfil}
\caption{\label{tab:guideline-example}
Example of entailment labels between source document (text) and headline (hypothesis). An italic part presents a paraphrase, and an underlined part presents a deviation.} 
\end{table*}

This section explores the hypothesis: {\it do source documents include sufficient information to produce headlines?}
We examine this hypothesis by considering textual entailment between a source document and its headline.
More specifically, we would like to know whether a source document entails its headline, i.e., whether we can infer that a headline is true based on the information in the source document.

We asked three human subjects to judge entailment relations for 1,000 pairs of source documents and headlines of each dataset.
We randomly selected 1,000 pairs from the test set of the English Gigaword dataset and 1,000 pairs from JAMUL.
The labels include {\it entail}, {\it non-entail}, and {\it other} (see Appendix for the definition of the labels and the treatment).

\begin{table}
\begin{center}
\begin{tabular}{lrrr}
\hline
\textbf{Dataset} & \textbf{Lead-1} & \textbf{Lead-3} & \textbf{Full}\\
\hline
Gigaword & 70.3\% & N/A & 92.8\% \\
JAMUL & N/A & 61.4\% & 94.2\% \\
\hline
\end{tabular}
\end{center}
\caption{\label{tab:entail-rate}
Ratio of document-headline pairs where the source documents entail their headlines.
}
\end{table}



Table~\ref{tab:entail-rate} reports the ratio of document-headline pairs for which two or three human subjects voted `yes' for the entailment relation ({\it entail}).
Only 70.3\% of lead-headline pairs in the Gigaword dataset hold the entailment relation.
For reference, we did the same analysis by using full-text articles as source documents and found that the ratio rises to 92.8\%. 
Similarly, only 61.4\% of lead three sentences (lead-3) and headline pairs in JAMUL hold the entailment relation.
When using full-text articles, the entailment ratio rises to 94.2\%.
These results support our hypothesis that source documents contain insufficient information under the current task settings.

\section{Improving the truthfulness of data}\label{sec:improve-data}

Based on the analysis in the previous section, we can consider two strategies to improve the task setting: using full-text articles as source documents instead of leading sentences; and removing non-entailment instances from the dataset.
Although the former strategy reduces the ratio of non-entailment pair to 7.2\% (English Gigaword) and 5.8\% (JAMUL), we must consider the trade-off: the use of full-text articles increases the cost for training, and may decrease the quality of headlines because of longer inputs to encoder-decoder models.
Furthermore, JNC does not provide full-text articles but only lead three sentences.
Therefore, we take the latter strategy, removing non-entailment pairs from the supervision data for headline generation.

\subsection{Recognizing textual entailment}\label{sec:rte-model}

In order to find non-entailment pairs in the dataset, we build a binary classifier that judges whether a source document entails its headline or not.
Recently, pretrained language models such as BERT~\cite{devlin-etal-2019-bert} show remarkable advances in the task of recognizing textual entailment (RTE)\footnote{\url{https://gluebenchmark.com/leaderboard}}.
Thus, we fine-tune pretrained models on the supervision data for entailment relation between source documents and their headlines.

For English Gigaword dataset, we use the pretrained RoBERTa large~\citep{liu2019roberta} fine-tuned on Multi-Genre Natural Language Inference (MultiNLI)~\citep{williams2018mnli}.
We further fine-tuned the model on the supervision data of the lead-headline pairs with entailment labels (acquired in Section~\ref{sec:analysis-dataset}).
Here, the supervision data include lead-headline pairs where two or three human subjects labeled either {\it entail} or {\it non-entail}; other pairs were excluded from the supervision data.
In this way, we obtained a binary classifier for entailment relation of 91.7\% accuracy on a hold-out evaluation (761 training and 179 test instances) after running 10 epoch of fine-tuning on the RoBERTa model.

For JNC, we use the pretrained BERT model for Japanese text~\citep{bertjapanese}.
However, no large-scale Japanese corpus for semantic inference (counterpart to MultiNLI) is available.
Thus, we created supervision data for entailment relation between lead three sentences and headlines ({\it lead3-headline}, hereafter) on JNC.
We extracted 12,000 lead3-headline pairs from JNC, and collected entailment labels using crowdsourcing.
Each pair had five entailment labels assigned by five crowd workers.
We used lead3-headline pairs where four or five crowd workers labeled either {\it entail} or {\it non-entail}; other pairs were unused in the supervision data.
The entailment classifier fine-tuned on the supervision data achieved 83.9\% accuracy on a hold-out evaluation with 5,033 training and 1,678 test instances.

Applying the entailment classifiers to the training and development sets of English Gigaword dataset and JNC, we removed instances of non-entailment pairs judged by the classifiers.
Eventually, we obtained 2,695,325 instances (71\% of the original training instances) on the English Gigaword dataset and 841,640 instances (49\% of the original training instances) on JNC.

\section{Improving the truthfulness of models}\label{sec:train}

In this section, we examine whether the supervision data built in the previous section reduces untruthful headlines.

\subsection{Headline generation models}

We use fairseq\footnote{\url{https://github.com/pytorch/fairseq}}~\citep{ott2019fairseq} as an implementation of the Transformer architecture~\citep{vaswani2017attention} throughout the experiments.
Hyper-parameter configurations are: 6 layers both in the encoder and decoder; 8 attention heads; the dimension of hidden states is 512; the dimension of hidden states of the feed forward network is 2048; the smoothing rate, dropout rate, and label smoothing were set to 0.1; Adam optimizer with $\beta = 0.98$, the learning rate of 0.0005, and 4,000 warm-up steps.

We train the Transformer models on the supervision data with and without non-entailment instances.
Because removing non-entailment instances decreases the number of training instances, we also apply the self-training strategy~\citep{Takemae2018NeuralHG} to obtain the same amount of training instances to the full supervision data.
More specifically, we generated headlines for the source documents discarded in Section \ref{sec:rte-model}, and added pairs of source documents and generated headlines as pseudo supervision data.
The experiments compare models trained on the full supervision data ({\it full}), the one filtered by the entailment classifier ({\it filtered}), and the one filtered but augmented by the self-training ({\it filtered+pseudo}).

\subsection{Data preparation}\label{sec:modified-gigaword}

The experiments use the same data split of training (3.8M instances), development (390k instances), and test (380k instances) sets to \citet{rush-etal-2015-neural}.
In this study, we used 10,000 instances for evaluation that were sampled from the test set and unused in the analysis in Section~\ref{sec:analysis-dataset}.
We do not apply any replace operations for the English Gigaword dataset: digit masking, rare word to UNK, and lower-casing.
The dataset is tokenized by WordPiece~\citep{Wu2016GooglesNM} with the same vocabulary used in UniLM.

Splitting JNC into 1.7M training and 3k development instances, we evaluate the model on the JAMUL dataset.
We use SentencePiece\footnote{\url{https://github.com/google/sentencepiece}}~\citep{kudo-richardson-2018-sentencepiece} for tokenization.

\subsection{Evaluation protocol}

We evaluate the quality of generated headlines by using full-length F1 ROUGE scores\footnote{ROUGE scores were computed by SumEval.We used MeCab~\citep{kudo-etal-2004-applying} for Japanese tokenization.}, following the previous work.
However, \newcite{kryscinski-etal-2019-neural} reported that ROUGE scores between system and reference summaries had only a weak correlation with human judgments.
Furthermore, we would like to confirm whether the filtering strategy can improve the truthfulness of the model.
Therefore, we also report the support score, the ratio of entailment relation between source documents and generated headlines measured by the entailment classifiers (explained in Section~\ref{sec:rte-model}), and human evaluation about the truthfulness.

\subsection{Results}

\begin{table*}
\center
\begin{tabular}{llrrrrrrr}
\hline
\textbf{Dataset} & \multicolumn{2}{l}{\textbf{Training data (amount)}} & \multicolumn{1}{c}{\textbf{R-1}} & \multicolumn{1}{c}{\textbf{R-2}} & \multicolumn{1}{c}{\textbf{R-L}} & \multicolumn{1}{c}{\textbf{Sup}} & \multicolumn{1}{c}{\textbf{Entail}} & \multicolumn{1}{c}{\textbf{Truthful}} \\
\hline
 & Full & (3.8 M) & 35.80 & 17.63 & 33.69 & 75.38 & 85.78\% & 77.06\% \\
Gigaword & Filtered & (2.7 M) & 35.24 & 17.29 & 33.14 & 77.61 & 91.50\% & --- \\
 & Filtered+pseudo & (3.8 M) & {\bf 35.85} & {\bf 17.94} & {\bf 33.72} & {\bf 79.91} & {\bf 93.56\%} & {\bf 85.32\%} \\
\hline
 & Full & (1.7 M) & {\bf 48.08} & {\bf 22.21} & {\bf 40.02} & 89.10 & 90.29\% & 89.91\%\\
JAMUL & Filtered & (0.8 M) & 46.08 & 20.81 & 38.07 & 90.14 & 95.67\% & --- \\
 & Filtered+pseudo & (1.7 M) & 45.62 & 20.55 & 38.10 & {\bf 90.65} & {\bf 96.26\%} & {\bf 92.66\%}\\
 \hline
\end{tabular}
\caption{\label{tab:rouge}
Results on the test set. We used F1 full-length ROUGE score: R-1 (ROUGE-1), R-2 (ROUGE-2), and R-L (ROUGE-L). ``Sup'' denotes support score. ``Entail'' presents the percentage of outputs to which the entailment classifier predicts the entailment relation (built in Section~\ref{sec:rte-model}). ``Truthful'' show the percentage of outputs to which a human subject judged as truthful headlines.
}
\end{table*}

\begin{figure}
    \center
    \includegraphics[width=\hsize]{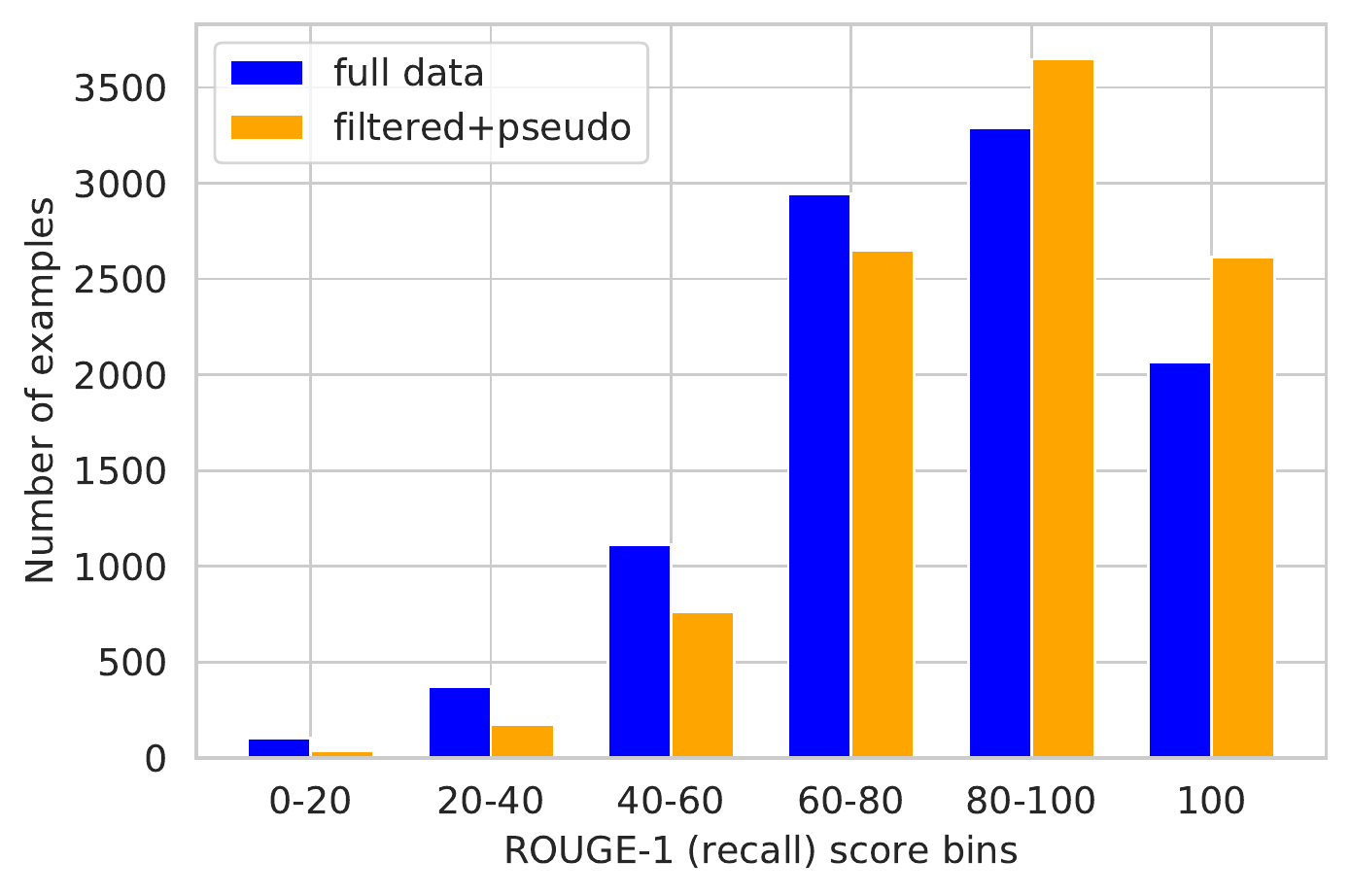}
    \caption{The distribution of the support scores on the English Gigaword dataset.}
    \label{fig:rouge-bins}
\end{figure}

Table~\ref{tab:rouge} shows the main results.
The baseline model with full training data obtained 35.80 ROUGE-1 score on the English Gigaword dataset and 48.08 ROUGE-1 score on JAMUL.
The entailment filter lowered ROUGE scores on both of the datasets probably because of the smaller number of training instances, but the self-training strategy improved ROUGE scores on the Gigaword dataset, outperforming the baseline model.

In contrast, the self-training strategy could not show an improvement for ROUGE scores on JAMUL.
Although it is difficult to find the exact cause of this result, we suspect that the filtering step reduced the training instances too much (0.8M instances) for the self-training method to be effective.
Another possibility is that the writing style of articles of non-entailment pairs in JNC/JAMUL is so distant that the self-training method generated headlines that are too different from reference ones.

\begin{figure}
    \center
    \includegraphics[width=\hsize]{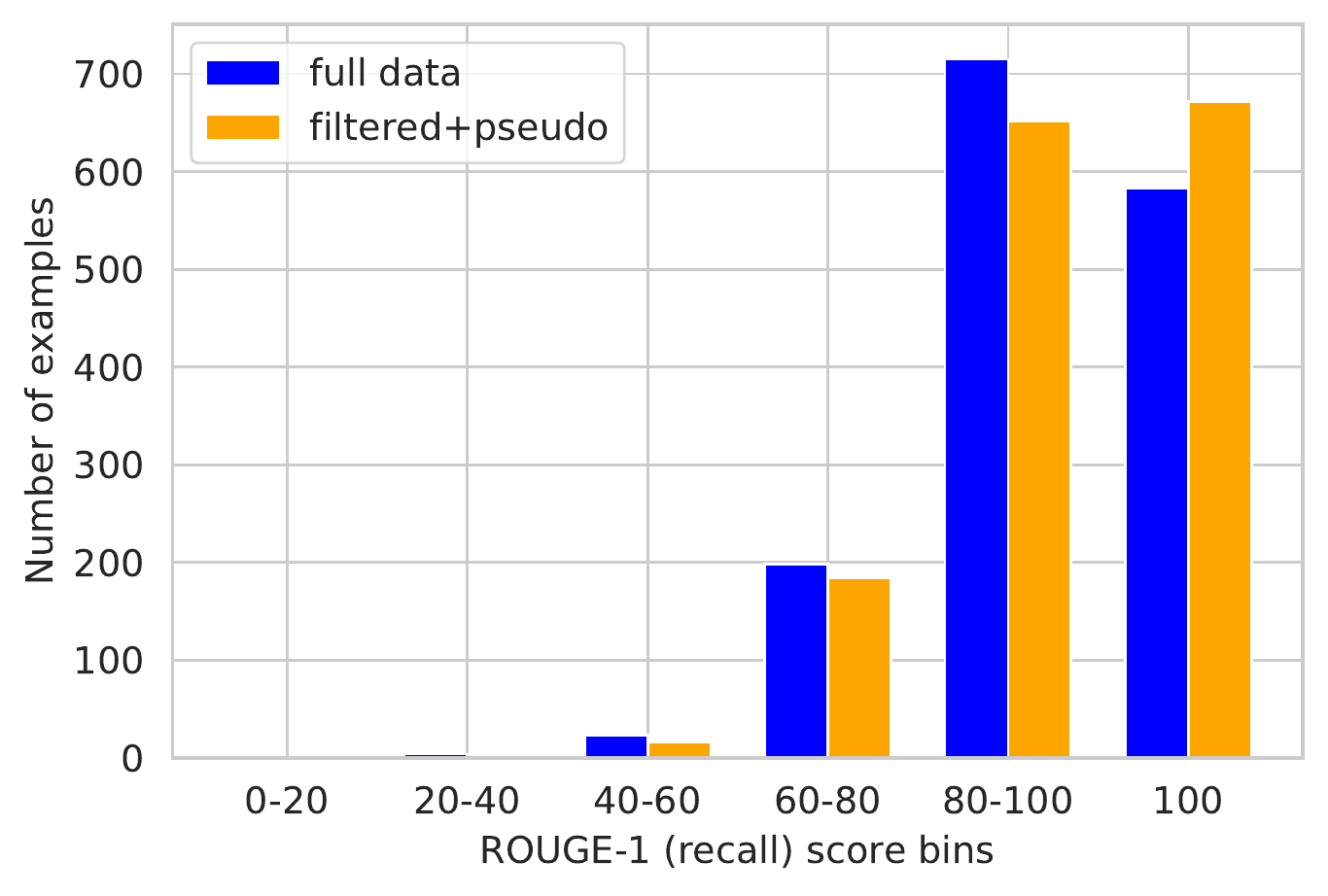}
    \caption{The distribution of the support scores on JAMUL.}
    \label{fig:rouge-bins-jamul}
\end{figure}

The column ``Sup'' presents the support score computed by the recall-oriented ROUGE-1 between source documents and generated headlines (explained in Section \ref{sec:estimating-truthfulness}).
The table indicates that the filtering and self-training strategies obtain higher support scores than the baseline.
Figures \ref{fig:rouge-bins} and \ref{fig:rouge-bins-jamul} depict histograms of the support scores for the baseline and filtering+pseudo settings on Gigaword and JAMUL, respectively.
We could confirm that the filtering+pseudo strategy increased the number of headlines with high support scores.

The column ``Entail'' shows the entailment ratio measured by the entailment classifier.
Again, the filtering+pseudo strategy obtained the highest entailment ratio on both the Gigaword dataset and JAMUL.
Although this result may be interpreted as natural because we selected training instances based on the same entailment classifier, it is interesting to see that we can control the entailment ratio without changing the model.

In order to examine whether the filtering strategy can deliver noticeable improvements for human readers, we asked a human subject to judge the truthfulness of the headlines generated by the baseline setting and filtering+pseudo strategy.
Presented with both a source document and a headline generated by the model, the human subject judged whether the headline was {\it truthful}, {\it untruthful}, or {\it incomprehensible}.
We conduct this evaluation for 109 instances randomly sampled from the test sets of Gigaword and JAMUL.

The ``Truthful'' column in Table~\ref{tab:rouge} reports the ratio of truthful headlines.
Consistently with the entailment ratio, we could confirm that the filtering+pseudo strategy generated truthful headlines more than the baseline setting on both of the datasets.
During the human evaluation, one instance in both full and filtered+pseudo settings from the Gigaword dataset judged as incomprehensible.


\subsection{Discussion}

To sum up the results, improving the truthfulness of the supervision data does help improving the truthfulness of generated headlines.
We could confirm the improvements from the support scores, entailment ratio, and human judgments.
However, the ROUGE scores between system and reference headlines did not indicate a clear difference.

The ROUGE metric was proposed to measure the {\it relevance} of a summary when extractive summarization was the central approach (in the early 2000s).
Obviously, the truthfulness of summaries is out of the scope of ROUGE.
The experimental results in this paper suggest that we should consider both {\it relevance} and {\it truthfulness} when evaluating the quality of abstractive summarization.

\section{Related Work}

\citet{rush-etal-2015-neural} first applied the neural sequence-to-sequence (seq2seq) architecture~\citep{Sutskever:2014:SSL, Bahdanau2014NeuralMT} to abstractive summarization.
They obtained a dataset for abstractive summarization from the English Gigaword~\citep{graff2003english,napoles-etal-2012-annotated}.
After this work, a large number of studies followed the task setting~\citep{takase-etal-2016-neural,zhou-etal-2017-selective,cao-etal-2018-retrieve,song2019mass,wang-etal-2019-biset}. 

Some researchers pointed out that abstractive summarization models based on seq2seq sometimes generate summaries with inaccurate facts. 
\citet{Cao2018FaithfulTT} reported that 30\% of the summaries generated by a seq2seq model include different facts from source articles.
In addition, \citet{kryscinski-etal-2019-neural} reported that ROUGE scores have only a weak correlation with human judgments in abstractive summarization and that the current evaluation protocol is inappropriate for factual consistency.

Several studies approach the problem of inconsistency between input and output by improving the model architecture or learning method.
\citet{Cao2018FaithfulTT} applied an information extraction tool to extract tuples of subject, predicate, and object from source documents and utilized them as an additional input to the model.
\citet{pasunuru2018multiRL} incorporated an entailment classifier as a reward in reinforcement learning. 
\citet{guo-etal-2018-soft} presented a multi-task learning method between summarization and entailment generation where hypotheses entailed by a given document (as a premise) are generated.
\citet{li-etal-2018-ensure} introduced an entailment-aware encoder-decoder model to ensure the correctness of the summary.
\citet{kiyono-etal-2018-reducing} reduced incorrect generations by modeling token-wise correspondences between input and output.
\citet{falke-etal-2019-ranking} proposed a re-ranking method of beam search based on factual correctness from a classifier of textual entailment.

As another direction, \citet{Kryscinski2019EvaluatingTF} evaluated the factual consistency of a source document and the generated summary with a weakly-supervised model.

A few studies raised concerns about the data set and task setting.
\citet{Tan:2017:NSS:3171837.3171860} argued that lead sentences do not provide an adequate source for the headline generation task.
The researchers reported that making use of multiple summaries as well as the lead sentence of an articles improved the performance of headline generation on the New York Times corpus.
In contrast, our paper is the first to analyze the truthfulness of  existing datasets and generated headlines, provide a remedy to the supervision data, and demonstrate the importance of truthfulness in headline generation.

\section{Conclusion and future work}

In this paper, we showed that the current headline generation model yields unexpected words.
We conjectured that one of the reasons lies in the defect in the task setting and data set, where generating a headline from the source document is impossible because of the insufficiency of the source information.
We presented an approach for removing from the supervision data headlines that are not entailed by their source documents.
Experimental results demonstrated that the headline generation model trained on filtered supervision data showed no clear difference in ROUGE scores but remarkable improvements in automatic and manual evaluations of the truthfulness of the generated headlines.
We also presented the importance of evaluating truthfulness in abstractive summarization.

In the future, we explore a more sophisticated method to improve the relevance and truthfulness of generated headlines, for example, removing only deviated spans in untruthful headlines rather than removing untruthful headlines entirely from the supervision data.
Other directions include an extensive evaluation of relevance and truthfulness of abstractive summarization and an establishment of an automatic evaluation metric for truthfulness.

Moreover, it will be also interesting to see whether the same issue occurs in other related tasks such as data-to-text generation.
We believe that the concern raised in this paper is beneficial to other tasks.


\section*{Acknowledgments}

The research results have been achieved by ``Research and Development of Deep Learning Technology for Advanced Multilingual Speech Translation'', the Commissioned Research of National Institute of Information and Communications Technology (NICT), Japan.

\bibliography{acl2020}
\bibliographystyle{acl_natbib}

\clearpage
\appendix

\section{Guideline for entailment labeling}
\label{sec:app:guideline}

Figure~\ref{fig:guideline} presents a guideline for the entailment labeling task in Section~\ref{sec:analysis-dataset}.
Given a pair of an article and headline, a crowd worker is expected to judge whether the article entails the headline, and label the pair with either of the labels shown in this figure.

\begin{figure}[t!]
\center
\framebox{\parbox{0.95\linewidth}{
\renewcommand{\labelenumi}{\arabic{enumi}.}
\textbf{Entail}
    　\begin{itemize}
      \item All facts of the headline are covered by those of the article.
      \item If the headline includes an expression that do not appear in the article, but if the fact mentioned by the expression can be derived from the article, judge the pair as ``Entail''.
      \end{itemize}
\textbf{Non-entail}
    　\begin{itemize}
      \item The statement of the headline conflicts with the article.
      \item The headline mentions facts that cannot be confirmed by the article.
      \end{itemize}
\textbf{Incomprehensible}
      \begin{itemize}
      \item Impossible to judge because the article or headline is unreadable. If the headline is not grammatically complete but correct as the headline style, please try to judge either entail or non-entail.
      \item Other problems such as garbled characters.
      \end{itemize}
}}

\caption{Guideline for entailment labeling}
\label{fig:guideline}
\end{figure}

\section{Examples}
\label{sec:app:examples}

\begin{figure}[t!]
\center
\begin{tabular}{p{6cm}}\hline
\textbf{Source:} \\
Suspected Muslim militants shot and killed five men who had formed a civilian patrol group to counter the radicals in their village , police officials said Monday . \\
\textbf{Full (baseline):} \\
Suspected Militants Kill Five \underline{in} \underline{Kashmir} \\
\textbf{Filtered+pseudo:} \\
Suspected Muslim Militants Kill Five \\ \hline

\textbf{Source:} \\
Divers searched the Mississippi River for bodies still trapped beneath the twisted debris of a collapsed freeway bridge Thursday , as finger - pointing began over a federal report two years ago that found the bridge was ` ` structurally deficient . \\
\textbf{Full (baseline):}\\
FEDERAL REPORT \underline{BEGINS} IN MISSISSIPPI \\
\textbf{Filtered+pseudo:} \\
Divers Search Mississippi River for Bodies \\ \hline

\textbf{Source:} \\
Tokyo stocks rose Tuesday as investors snapped up domestic demand - related issues due to receding jitters among investors over last week ' s plunge . \\
\textbf{Full (baseline):} \\
Tokyo stocks rise , \underline{dollar lower against} \underline{yen} \\
\textbf{Filtered+pseudo:} \\
Tokyo stocks end higher \\ \hline

\end{tabular}
\caption{\label{tab:improved-example}
Examples of the improved headlines.
} 
\end{figure}

Figure \ref{tab:improved-example} shows some examples of the generated headlines from the models described in Section \ref{sec:train}.
In the first example, the baseline model added ``in Kashmir'' in the headline, but this is incorrect. The correct location is in Southern Egypt, which was mentioned in the reference headline.
The filtered+pseudo model generates a safe headline.
The second headline generated by the baseline includes the verb `begin' although the report was written two years ago.
The baseline model added ``dollar lower against yen'' in the headline. There is a correlation indeed that dollar is lower against yen when Tokyo stocks rise, but we cannot confirm the fact from the source document.

\end{document}